\documentclass{article}

\usepackage[nonatbib,final]{neurips_2023}

\usepackage[utf8]{inputenc} 
\usepackage[T1]{fontenc}    
\usepackage{hyperref}       
\usepackage{url}            
\usepackage{booktabs}       
\usepackage{amsfonts}       
\usepackage{nicefrac}       
\usepackage{microtype}      
\usepackage[table]{xcolor}         
\usepackage[most]{tcolorbox}
\usepackage{mdframed}
\usepackage{array}
\usepackage[utf8]{inputenc} 
\usepackage[T1]{fontenc}    
\usepackage{hyperref}
\usepackage{url}
\usepackage{booktabs}
\usepackage{amsfonts}
\usepackage{nicefrac}
\usepackage{microtype}
\usepackage[export]{adjustbox}
\usepackage{amsmath}
\usepackage{tikz}
\usetikzlibrary{shadows.blur}
\usepackage[most]{tcolorbox}
\usepackage{caption}
\usepackage{array}
\usepackage{soul}
\usepackage[super]{nth}
\usepackage{float}
\usepackage{wrapfig}
\usepackage{mdframed}
\definecolor{lightgray}{rgb}{0.93, 0.93, 0.93}
\newrobustcmd{\B}{\bfseries}

\title{Platypus: Quick, Cheap, and Powerful \\ Refinement of LLMs}
\author{
  Ariel N. Lee\footnotemark[1] \\
  Boston University\\
  \texttt{ariellee@bu.edu} \\
  \And
  Cole J. Hunter\footnotemark[1] \\
  Boston University\\
  \texttt{colejh@bu.edu}  \\
  \And
  Nataniel Ruiz\footnotemark[2] \\
  Boston University\\
  \texttt{nruiz9@bu.edu} \\
}
\begin{document}

\maketitle

\footnotetext[1]{Equal Contribution.}
\footnotetext[2]{NR is currently at Google and his contributions were done as work at BU prior to his tenure at the company.}

\begin{center}
\centering
\begin{tikzpicture}
\centering
    \clip (0,0) circle (3cm);
    \node at (0,0) {\includegraphics[width=6cm]{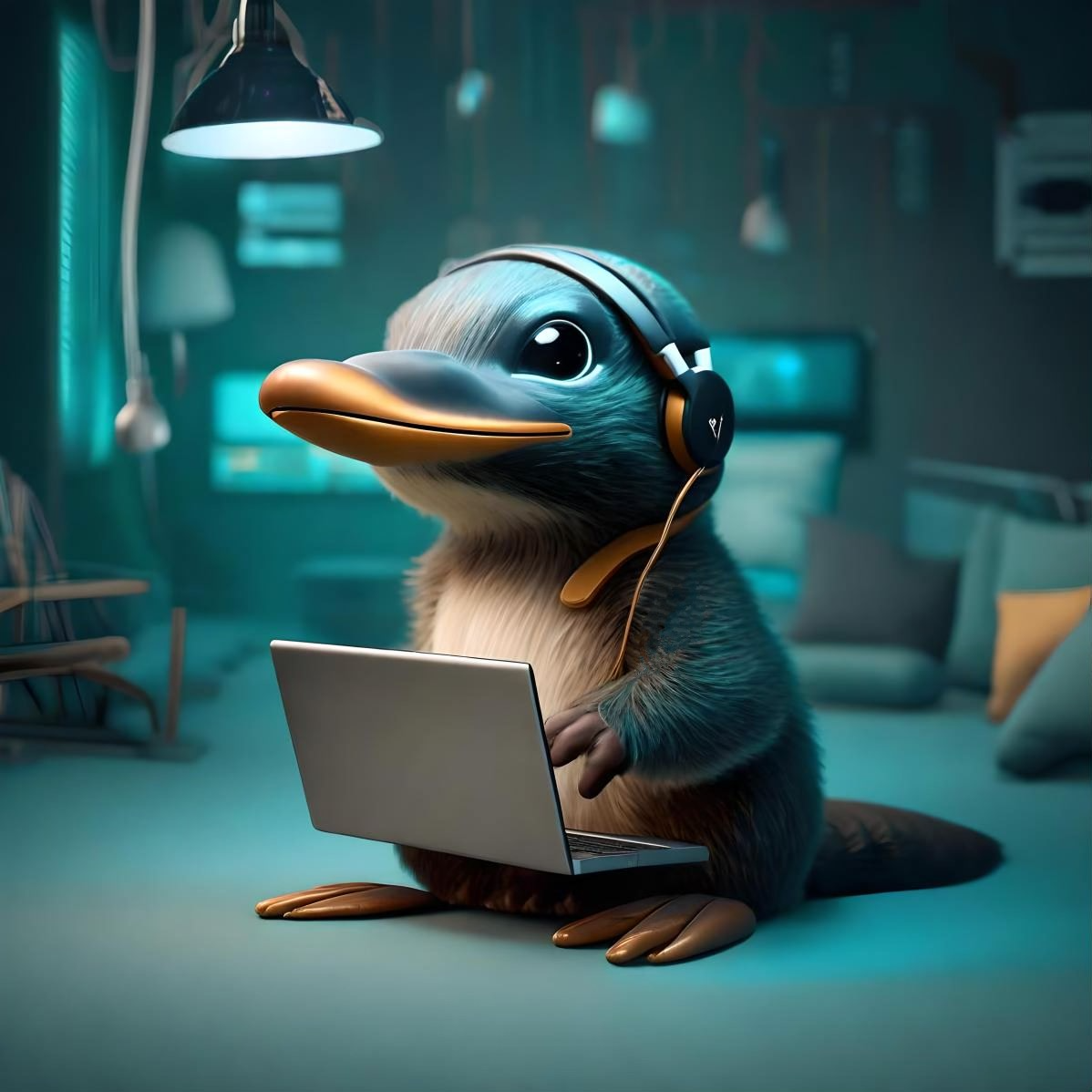}};
\end{tikzpicture}
\label{fig:teaser}
\end{center}

\vspace{10pt}

\begin{abstract}
We present \textbf{Platypus}, a family of fine-tuned and merged Large Language Models (LLMs) that achieved the strongest performance and stood at first place in HuggingFace's Open LLM Leaderboard~\footnote[1]{\url{https://huggingface.co/spaces/HuggingFaceH4/open_llm_leaderboard}} at the time of writing. In this work we describe (1) our curated dataset \textbf{Open-Platypus}, that is a subset of other open datasets and which \textit{we release to the public} (2) our process of fine-tuning and merging LoRA modules in order to conserve the strong prior of pretrained LLMs, while bringing specific domain knowledge to the surface (3) our efforts in checking for test data leaks and contamination in the training data, which can inform future research. Specifically, the Platypus family achieves strong performance in quantitative LLM metrics across model sizes, topping the global Open LLM leaderboard while using just a fraction of the fine-tuning data and overall compute that are required for other state-of-the-art fine-tuned LLMs. In particular, a 13B Platypus model can be trained on \textit{a single} A100 GPU using 25k questions in 5 hours. This is a testament of the quality of our Open-Platypus dataset, and opens opportunities for more improvements in the field. 
Project page: {\url{https://platypus-llm.github.io}}
\end{abstract}

\section{Introduction}
Our research centers around improving the performance of base Large Language Models (LLMs) by fine-tuning models using parameter efficient tuning (PEFT) on a small, yet powerful, curated dataset \textbf{Open-Platypus}. This work lives in the context of recent advancements in the domain of LLMs. The rapid growth of these models was kick-started by the emergence of scaling laws~\cite{kaplan2020scaling}. Soon after, 100B+ parameter models like PaLM~\cite{chowdhery2022palm} and GPT-3~\cite{brown2020language} were proposed. Task specific models came next, such as Galactica for scientific tasks~\cite{taylor2022galactica}. Chinchillia~\cite{hoffmann2022training} was introduced along with a novel scaling law approach that shifts the emphasis from model size to the number of processed tokens. 

To challenge the dominance of closed source models like OpenAI's GPT-3.5 and GPT-4, Meta released the original LLaMa models~\cite{touvron2023llama}, now known for their computational efficiency during inference. Open-source initiatives such as BLOOM~\cite{lescao2022bloom} and Falcon~\cite{almazrouei2023falcon} have also been released to challenge the hegemony of their closed-source counterparts. Recently, Meta AI released LLaMa-2 models~\cite{touvron2023llama2}. Shortly after the initial release the 70B parameter model was fine-tuned by StabilityAI to create StableBeluga2~\cite{StableBelugaModels} using an Orca-style dataset~\cite{mukherjee2023orca}. As the the scale of both network architectures and training datasets have grown, the push towards employing LLMs as generalist tools able to handle a wide array of tasks has intensified. For the largest models, their abilities as generalists make them well-suited for many NLP tasks~\cite{qin2023chatgpt}, with smaller models struggling to maintain the same level of versatility. 

A number of strategies have been employed to try and bridge this divide. A prominent method known as knowledge distillation~\cite{hsieh2023distilling,hinton2015distilling,lamini-lm} aims to transfer knowledge from a large, more performant teacher model to a smaller student model, preserving performance while reducing computational overhead. Recently, the most popular method involves distilling the knowledge from a large training dataset into a small one, again making it less computationally expensive than traditional approaches~\cite{yu2023dataset}. These methods also tend to take advantage of \textit{instruction tuning}~\cite{wei2022finetuned}, which has proven an effective method for improving the general performance of LLMs. Projects like Stanford's Alpaca~\cite{alpaca} and WizardLM~\cite{xu2023wizardlm} provide frameworks for generating high-quality, instruction formatted data. Fine-tuning base models on these types of datasets and applying self-instruct methodology~\cite{selfinstruct} has led to marked improvements in both their quantitative and qualitative performance~\cite{chung2022scaling}.

The Mixture of Experts approach~\cite{shen2023mixtureofexperts, shazeer2017outrageously} employs conditional computation, activating network sections based on individual examples. This technique boosts model capacity without a linear rise in computation. Sparse variants, like the Switch Transformer~\cite{fedus2022switch}, activate select experts per token or example, introducing network sparsity. Such models excel in scalability across domains and retention in continual learning, as seen with Expert Gate~\cite{expertgate}. Yet, ineffective expert routing can result in under-training and uneven specialization of experts.

Following the recent arrival of LoRA is Quantized\-LoRA (QLoRA)~\cite{dettmers2023qlora}, which has been recognized as an efficient and cost-effective methodology. The authors of~\cite{dettmers2023qlora} concurrently released Guanaco, a new model family. The best Guanaco models currently rank \nth{7} and \nth{12} on the Hugging Face leaderboard as of the time of writing. Notwithstanding, our initial decision to employ LoRA occurred before the release of QLoRA, and we stuck with it since it proved effective within our existing workflow—namely being compatible and successful at model merging. Since our future goals include reducing training time and cost, we would be excited to use quantized LoRA in our pipeline and compare results.

Other approaches have centered on training LLMs in specific tasks such as coding~\cite{luo2023wizardcoder}, quantitative reasoning~\cite{lewkowycz2022solving}, and biomedical knowledge~\cite{singhal2023expertlevel}. This specialized training has its own merits. By focusing on narrower domains, these models can achieve higher accuracy rates and more relevant output in their respective fields. 

One \textit{large limitation} of this approach, especially for domain-specific models derived from large, pre-trained ones, is that the fine-tuning process can be \textbf{time-consuming} and \textbf{costly}. Our work seeks to address these issues by focusing on refining a training recipe aimed to maintain the benefits of instruction tuning, namely \textit{generalized improvement}, while also imparting \textit{specific domain knowledge}. We find that domain specific datasets increase performance on a selected category of tasks, which when combined with merging significantly reduces training time.
Our core contributions are as follows:
\begin{itemize}
    \item \textbf{Open-Platypus}, a small-scale dataset that consists of a curated sub-selection of public text datasets. The dataset is focused on improving LLMs' STEM and logic knowledge, and is made up of 11 open-source datasets. It is comprised mainly of human-designed questions, with only \(\sim \)10\% of questions generated by an LLM. The main advantage of Open-Platypus is that, given its size and quality, it allows for very strong performance with short and cheap fine-tuning time and cost. Specifically, one can train their own 13B model on a single A100 GPU using 25k questions in 5 hours.
    \item A description of our process of similarity exclusion in order to reduce the size of our dataset, as well as reduce data redundancy.
    \item A detailed look into the ever-present phenomenon of contamination of open LLM training sets with data contained in important LLM test sets, and a description of our training data filtering process in order to avoid this pitfall. 
    \item A description of our selection and merging process for our specialized fine-tuned LoRA modules.
\end{itemize}

\section{Methods}

\subsection{Curating Open-Platypus}
Our decisions regarding data selection for fine-tuning the LLaMa-2 models were influenced by (1) the Superficial Alignment Hypothesis presented by~\cite{zhou2023lima}, which states that model knowledge is almost entirely learned during pre-training, and that with minimal training data it is possible to achieve excellent results aligning model outputs; (2) the LLaMa2 introductory paper in which~\cite{touvron2023llama2} state that the base models had not yet reached saturation; and (3) the work of~\cite{gunasekar2023textbooks}, highlighting the importance of high-quality input data for training effective models. Put into practice, and keeping in mind our goal of optimizing training time and model performance, our approach to fine-tuning the LLaMa-2 models was a balanced blend of the three points above. By focusing on depth in specific areas, diversity of input prompts, and keeping the size of the training set small, we aimed to maximize the precision and relevance of our models' outputs. To achieve this, we curated a content filtered, instruction tuned dataset which draws from a variety of open-source datasets.  In this context, 'content filtered' refers to our choice for the train set to almost exclusively include data which is related to our domain of interest, namely STEM. 

Open-Platypus is made up of 11 open-source datasets, detailed in Table \ref{tab:datasets}. It is comprised mainly of human-designed questions, with \(\sim \)10\% of questions generated by an LLM.  Given our focus on STEM and logic, we primarily pulled from datasets geared towards those subjects, supplementing them with keyword-filtered content from datasets with a broader subject coverage, namely Openassistant-Guanaco~\cite{dettmers2023qlora} and airoboros~\cite{jondurbin2023airoboros}. The backbone of Open-Platypus is a modified version of MATH~\cite{hendrycksmath2021} that has been supplemented with expanded step-by-step solutions from PRM800K~\cite{lightman2023lets}.

\begin{table}[ht]
     \small
     \caption{Datasets, Licenses, and Number of Leaked Questions. With respect to Open-Platypus, after using keyword searches to filter for STEM and logic, we \textit{removed any training questions} with similarity > 80\% to any test set question. \textbf{*}The datasets marked with asterisks were not added to Open-Platypus but we include them because we ran contamination checks when considering which models to merge.}
     \vspace{1\baselineskip}
     \label{tab:datasets}
     \centering
     \rowcolors{2}{white}{lightgray}
     \begin{tabular}{>{\raggedright\arraybackslash}p{4.7cm} >{\raggedright\arraybackslash}p{4.7cm} p{3cm}}
     \toprule 
     \textbf{Dataset Name} & \textbf{License Type} & \textbf{\# Leaked Questions} \\
     \midrule
     PRM800K: A Process Supervision Dataset~\cite{lightman2023lets} & MIT & 77 \\
     Measuring Mathematical Problem Solving With the MATH Dataset~\cite{hendrycksmath2021} & MIT & 77 \\
     ScienceQA: Science Question Answering~\cite{lu2022learn} & Creative Commons Attribution-NonCommercial-ShareAlike 4.0 & 0 \\
     SciBench: Evaluating College-Level Scientific Problem-Solving Abilities of Large Language Models~\cite{wang2023scibench} & MIT & 0 \\
     ReClor: A Reading Comprehension Dataset Requiring Logical Reasoning\cite{yu2020reclor} & Non-commercial & 0 \\
     *SciQ: Crowdsourcing Multiple Choice Science Questions~\cite{SciQ} & Creative Commons Attribution-NonCommercial 3.0 & 71 \\
     TheoremQA: A Theorem-driven Question Answering Dataset~\cite{chen2023theoremqa} & MIT & 0 \\
     \texttt{leetcode-solutions-python -testgen-gpt4}~\cite{leetcodesolutionsgpt4} & None listed & 0 \\
     \texttt{airoboros-gpt4-1.4.1}~\cite{jondurbin2023airoboros} & other & 13 \\
     \texttt{tigerbot-kaggle -leetcodesolutions-en-2k}\cite{leetcodesolutionskaggle} & apache-2.0 & 0 \\
     ARB: Advanced Reasoning Benchmark for Large Language Models~\cite{sawada2023arb} & MIT & 0 \\
     Openassistant-guanaco~\cite{dettmers2023qlora} & apache-2.0 & 13 \\
     *\texttt{ehartford/dolphin} (first 25k rows)~\cite{dolphin} & apache-2.0 & 0 \\
     \bottomrule
     \end{tabular}
 \end{table}

We employed the Alpaca instruction-tuning format, wherein each question is structured with an instruction, input, and output. In many cases the input is empty. However, for some datasets consisting of multiple choice questions, specifically ARB~\cite{sawada2023arb} and ReClor~\cite{yu2020reclor}, we integrated the formatting context \{\texttt{Choose A, B, C, or D}\} as input for each question. For ScienceQA~\cite{lu2022learn}, we opted to include long-form answers to the multiple choice questions, omitting an explicit statement of the correct choice entirely.

\subsection{Removing similar \& duplicate questions}
Having collected data from a number of sources, we then ran it through a de-duplication process to minimize the chances of memorization~\cite{lee2022deduplicating}. First, we removed all instructions which were word-for-word duplicates, followed by removal of instructions which had 80\% cosine similarity with the SentenceTransformers~\cite{reimers-2019-sentence-bert} embeddings of other instructions in our train set. In both cases, we defaulted to keeping the question-answer pair which had the more verbose answer. Our motivation behind this was that longer answers likely translate to more detailed explanations and/or step-by-step solutions.

\subsection{Contamination Check}
A core component of our methodology revolves around ensuring that none of the benchmark test questions inadvertently leak into the training set, which is a fairly common occurrence. We seek to try and prevent memorization of test data skewing the benchmark results. With that in mind, we did allow for some leniency in determining whether questions should be marked as duplicates and removed from the training set. Allowing some flexibility in identifying suspect questions acknowledges that there are multiple ways to phrase a query, and general domain knowledge might prevent a question from being considered duplicate.

To that end, we developed the following heuristics to guide manual filtering of questions from Open-Platypus that scored > 80\% similarity to any benchmark questions. We categorize potential leaks into three groups: duplicate, gray-area, and similar but different. For our purposes, we err on the side of caution and remove all of them from our train set.

\begin{figure}[ht]
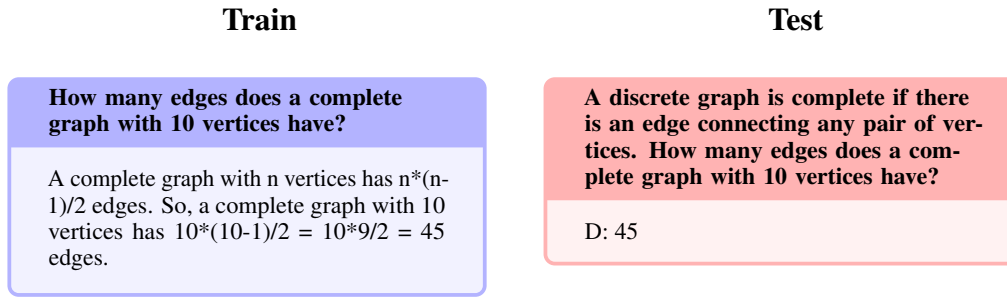

    \centering
    \small
    \begin{tabular}{cc}
    \multicolumn{1}{c}{\textbf{\large{Train}}} & \multicolumn{1}{c}{\textbf{\large{Test}}} \\
        \begin{minipage}[t]{0.48\textwidth}\rule{0pt}{0pt}
            \begin{tcolorbox}[colback=blue!5, colframe=blue!30, coltitle=black, drop shadow={blur shadow}, width=0.95\textwidth, rounded corners, title=\textbf{How many edges does a complete graph with 10 vertices have?}, halign title=left]
            A complete graph with n vertices has n*(n-1)/2 edges. So, a complete graph with 10 vertices has 10*(10-1)/2 = 10*9/2 = 45 edges.
            \end{tcolorbox}
        \end{minipage} &
        \begin{minipage}[t]{0.48\textwidth}\rule{0pt}{0pt}
            \begin{tcolorbox}[colback=red!5, colframe=red!30, coltitle=black, drop shadow={blur shadow}, width=0.95\textwidth, rounded corners, title=\textbf{A discrete graph is complete if there is an edge connecting any pair of vertices. How many edges does a complete graph with 10 vertices have?}, halign title=left]
             D: 45
            \end{tcolorbox}
        \end{minipage}
    \end{tabular}
    \caption{Comparison of train and test questions in the duplicate contamination group.}
    \label{fig:dups}
\end{figure}

\paragraph{Duplicate}
Questions marked as duplicate contamination are essentially exact copies of questions found in the test sets. This includes training questions with an extra word or minor rearrangement in relation to a benchmark question. Duplicate contamination is the only category we count as "true" contamination and corresponds to the number of leaked questions listed in Table \ref{tab:datasets}. Specific examples of this can be seen in Figure \ref{fig:dups}.

\begin{figure}[ht]
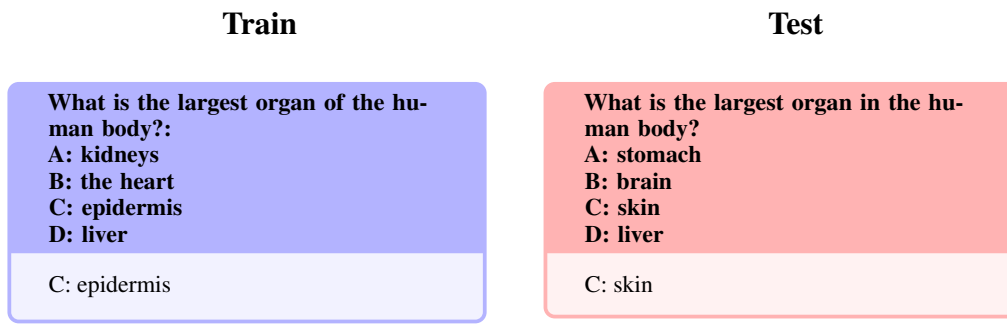

    \centering
    \small
    \begin{tabular}{cc}
    \multicolumn{1}{c}{\textbf{\large{Train}}} & \multicolumn{1}{c}{\textbf{\large{Test}}} \\
        \begin{minipage}[t]{0.48\textwidth}\rule{0pt}{0pt}
            \begin{tcolorbox}[colback=blue!5, colframe=blue!30, coltitle=black, drop shadow={blur shadow}, width=0.95\textwidth, rounded corners, title=\textbf{What is the largest organ of the human body?: \\
            A: kidneys \\
            B: the heart \\
            C: epidermis \\
            D: liver}, halign title=left]
            C: epidermis
            \end{tcolorbox}
        \end{minipage} &
        \begin{minipage}[t]{0.48\textwidth}\rule{0pt}{0pt}
            \begin{tcolorbox}[colback=red!5, colframe=red!30, coltitle=black, drop shadow={blur shadow}, width=0.95\textwidth, rounded corners, title=\textbf{What is the largest organ in the human body? \\
            A: stomach \\
            B: brain \\
            C: skin \\
            D: liver}, halign title=left]
            C: skin
            \end{tcolorbox}
        \end{minipage} \\
    \end{tabular}
    \caption{Comparison of train and test questions in the gray-area.}
    \label{fig:gray}
\end{figure}

\paragraph{Gray-area}
The next group, termed gray-area, encompasses questions that are not exact duplicates and fall within the realm of general knowledge. While we leave the final judgement of these questions to the open-source community, we believe they often necessitate expert knowledge. Notably, this category includes questions with identical instructions but answers that are synonymous (for instance, "skin" versus "epidermis" as illustrated in \ref{fig:gray}). It also covers questions that, though phrased differently, have their precise or detailed answers present in the training output. 

\paragraph{Similar but different}
The final category includes questions that, despite having high cosine similarity scores, yield very different answers. This can often be attributed to subtle changes in the question's structure, resulting in completely distinct responses. A great example of this is the question in Figure \ref{fig:simdiff}, where modifications to the rotation axis and quadrant definition significantly alter the final answer. 

\begin{figure}[ht]
    \centering
    \small
    \begin{tabular}{cc}
    \multicolumn{1}{c}{\textbf{\large{Train}}} & \multicolumn{1}{c}{\textbf{\large{Test}}} \\
        \begin{minipage}[t]{0.48\textwidth}\rule{0pt}{0pt}
            \begin{tcolorbox}[colback=blue!5, colframe=blue!30, coltitle=black, drop shadow={blur shadow}, width=0.95\textwidth, rounded corners, title=\textbf{The region \texttt{\$\textbackslash mathscr\{R\}\$} enclosed by the curves \texttt{\$y=x\$} and \texttt{\$y=x\texttt{\char`\^}2\$} is rotated about the \texttt{\$x\$}-axis. Find the volume of the resulting solid.}, halign title=left]
            The curves \texttt{\$y=x\$} and \texttt{\$y=x\texttt{\char`\^}2\$} intersect at the points \texttt{\$(0,0)\$} and \texttt{\$(1,1)\$}. The region between them, the solid of rotation, and a cross-section perpendicular to the \texttt{\$x\$}-axis are shown in Figure. A cross-section in the plane \texttt{\$P\_x\$} has the shape of a washer (an annular ring) with inner radius \texttt{\$x\texttt{\char`\^}2\$} and outer radius \texttt{\$x\$}, so we find the cross-sectional area by subtracting the area of the inner circle from the area of the outer circle: \texttt{\textbackslash r\textbackslash n\$\$\textbackslash r\textbackslash nA(x)=\textbackslash\textbackslash pi x\texttt{\char`\^}2-\textbackslash\textbackslash pi \textbackslash\textbackslash left(x\texttt{\char`\^}2\textbackslash\textbackslash right)\texttt{\char`\^}2= \textbackslash\textbackslash pi \textbackslash\textbackslash left(x\texttt{\char`\^}2-x\texttt{\char`\^}4\textbackslash\textbackslash right)\textbackslash r\textbackslash n\$\$\textbackslash r \textbackslash n} Therefore we have \texttt{\textbackslash r\textbackslash n\$\$ \textbackslash r\textbackslash n\textbackslash\textbackslash begin\{aligned\} \textbackslash r\textbackslash nV \& =\textbackslash\textbackslash int\_0\texttt{\char`\^}1 A(x) d x=\textbackslash\textbackslash int\_0\texttt{\char`\^}1 \textbackslash\textbackslash pi\textbackslash\textbackslash left(x\texttt{\char`\^}2-x\texttt{\char`\^}4\textbackslash\textbackslash right) d x \textbackslash\textbackslash\textbackslash\textbackslash\textbackslash r\textbackslash n\& =\textbackslash\textbackslash pi \textbackslash\textbackslash left[\textbackslash\textbackslash frac\{x\texttt{\char`\^}3\}\{3\} - \textbackslash\textbackslash frac\{x\texttt{\char`\^}5\}\{5\} \textbackslash\textbackslash right]\_0\texttt{\char`\^}1 = \textbackslash\textbackslash frac\{2 \textbackslash\textbackslash pi\}\{15\}\textbackslash r\textbackslash n\textbackslash\textbackslash end\{aligned\}\textbackslash r\textbackslash n\$\$}
            \end{tcolorbox}
        \end{minipage} &
        \begin{minipage}[t]{0.48\textwidth}\rule{0pt}{0pt}
            \begin{tcolorbox}[colback=red!5, colframe=red!30, coltitle=black, drop shadow={blur shadow}, width=0.95\textwidth, rounded corners, title=\textbf{The region bounded by the curves y = x and y = x\texttt{\char`\^}2 in the first quadrant of the xy-plane is rotated about the y-axis. The volume of the resulting solid of revolution is}, halign title=left]
            B: pi / 6
            \end{tcolorbox}
        \end{minipage} \\
    \end{tabular}
    \captionsetup{justification=centering}
    \caption{Comparison of train and test questions with high cosine similarity scores which are actually quite different.}
    \label{fig:simdiff}
\end{figure}

\subsection{Fine-tuning \& merging}
After refining the dataset and triple checking for contamination, our methodology centers on two main points: the effectiveness of Low Rank Approximation (LoRA)~\cite{hu2022lora} training and the built-in model merging capabilities of the State-of-the-art Parameter-Efficient Fine-Tuning (PEFT) library~\cite{peft}. Different from full fine-tuning methods, LoRA freezes pre-trained model weights and adds rank decomposition matrices into each layer of the transformer. This reduces the number of trainable parameters for downstream tasks and by extension, the time and cost of training. For example, our 13B model was fine-tuned using 1 A100 80GB for 5 hours and our 70B model using 4 A100s 80GB for 22 hours. As a benchmark for comparison, Stanford notes that their full fine-tune of Alpaca-7B took 3 hours on 8 A100s 80GB. In addition to PEFT and LoRA, we fine-tuned our models using the Hugging Face transformers library~\cite{HF-transformers}. As previously mentioned, we utilized Stanford Alpaca's prompt formatting template~\cite{alpaca}, which can found in the Appendix.

Our initial attempts at fine-tuning the models focused on the attention modules \texttt{v\_proj}, \texttt{q\_proj}, \texttt{k\_proj}, and \texttt{o\_proj}. We later moved onto the \texttt{gate\_proj}, \texttt{down\_proj}, and \texttt{up\_proj} modules as recommended by~\cite{he2022unified}, due to their analysis showing superior performance compared to the attention modules, with the exception being situations where the trainable parameters are a tiny fraction ($< 0.1\%$) of total parameters. For consistency, we adopted this strategy for both the 13 and 70 billion parameter fine-tunes, which translated to 0.27\% and 0.2\% trainable parameters, respectively. Please see the full list of hyperparameters in Table \ref{tab:hyperparams}. The only difference between our 13B and 70B models is the initial learning rate—we had to lower the initial learning rate for the 70B model from 4e-4 to 3e-4 because the loss went to zero after 15 steps. LoRA rank defines the dimensions of the low-rank matrices, and LoRA alpha is the scaling factor for the weight matrices. The weight matrix is scaled by $\frac{lora\_alpha}{lora\_rank}$, and a higher alpha value assigns more weight to the LoRA activations. We chose 16 since this was common practice in training scripts we reviewed and chose a 1:1 ratio so as not to overpower the base model.

After reviewing the datasets in Table \ref{tab:datasets}, we deliberately chose not to merge with any models trained using contaminated datasets. For example, we merged with the new Dolphin-70B LLM after confirming no test questions had leaked into the training set. We performed contamination checks on datasets used to train models we merged with to the best of our abilities, but some datasets have not been publicly released. While we cannot offer absolute assurances for any merged models with closed-source datasets, we proceed giving the benefit of the doubt. Additional details regarding merging considerations are included in the next section, as this is dependent on the fine-tune benchmark results. 

\section{Results}

In this section, we present a detailed analysis of our models' performance, bench-marking them against other state-of-the-art models. Our primary objective was to discern the effects of merging both broad and niche models and to assess the advantages of fine-tuning on our dataset. Moving forward, base model refers to the model on which the LoRA adapters are merged. 

Per the Hugging Face Open LLM Leaderboard data from 8/10/23 (Table \ref{tab:HF}), our Platypus2-70B-instruct variant has outperformed its competitors, securing the top position with an average score of 73.13. Notably, our Stable-Platypus2-13B model, as shown in Table \ref{tab:13B_summary}, stands out as the premier 13 billion parameter model with an average score of 63.96.

\begin{table}[ht]
    \centering
    \rowcolors{2}{white}{lightgray}
    \vspace{1\baselineskip}
    \caption{Top 15 open-source models on 8/10/23, including GPT-4 and GPT-3.5, according to the Hugging Face Open LLM leaderboard. Please note that GPT-4 and GPT-3.5 are not part of the official leaderboard but we have added their benchmark results for a closed-source model comparison. Our models are in \nth{1}, \nth{5}, \nth{11}, and \nth{15}. ARC-challenge is 25-shot, HellaSwag is 10-shot, MMLU is 5-shot, and TruthfulQA is 0-shot. \textbf{*}Note: Camel-Platypus2-70B is currently pending evaluation on the leaderboard, so we have included our local benchmark results instead.}
    \label{tab:HF}
    \setlength\tabcolsep{0.5\tabcolsep}
    \vspace{1\baselineskip}
    \begin{tabular}{@{}p{6.4cm}@{}ccccc}
    \toprule 
    \textbf{Model} & \textbf{Avg.} & \multicolumn{4}{c}{\textbf{Scores (\%)}} \\
    \cmidrule(lr){3-6}
    & & \textbf{ARC} & \textbf{HellaSwag} & \textbf{MMLU} & \textbf{TruthfulQA} \\
    \midrule
    \raggedright gpt-4 & 84.3 & 96.3 & 95.3 & 86.4 & 59 \\
    \midrule
    \raggedright \textbf{1. Platypus2-70B-instruct} & 73.13 & 71.84 & 87.94 & 70.48 & 62.26 \\
    \raggedright 2. upstage/Llama-2-70b-instruct-v2 & 72.95 & 71.08 & 87.89 & 70.58 & 62.25 \\
    \raggedright 3. psmathur/model\_007 & 72.72 & 71.08 & 87.65 & 69.04 & 63.12 \\
    \raggedright 4. upstage/Llama-2-70b-instruct & 72.29 & 70.9 & 87.48 & 69.8 & 60.97 \\
    \midrule
    \raggedright gpt-3.5 & 71.9 & 85.2 & 85.5 & 70 & 47 \\
    \midrule
    \raggedright \textbf{5. *Camel-Platypus2-70B} & 71.60 & 71.16 & 87.66 & 69.80 & 57.77 \\
    \raggedright 6. stabilityai/StableBeluga2 & 71.42 & 71.08 & 86.37 & 68.79 & 59.44 \\
    \raggedright 7. quantumaikr/llama-2-70b-fb16 -guanaco-1k & 71.41 & 70.48 & 87.33 & 70.25 & 57.56 \\
    \raggedright 8. augtoma/qCammel-70-x & 70.97 & 68.34 & 87.87 & 70.18 & 57.47 \\
    \raggedright 9. jondurbin/airoboros-l2-70b-gpt4-1.4.1 & 70.93 & 70.39 & 87.82 & 70.31 & 55.2 \\
    \raggedright 10. dfurman/llama-2-70b-dolphin-peft & 70.76 & 69.62 & 86.82 & 69.18 & 57.43 \\
    \raggedright \textbf{11. Dolphin-Platypus2-70B} & 70.69 & 70.39 & 86.7 & 69.04 & 56.65 \\
    \raggedright 12. TheBloke/llama-2-70b-Guanaco-QLoRA-fp16 & 70.63 & 68.26 & 88.32 & 70.23 & 55.69 \\
    \raggedright 13. psmathur/model\_420 & 70.55 & 70.14 & 87.73 & 70.35 & 54 \\
    \raggedright 14. psmathur/model\_51 & 70.41 & 68.43 & 86.71 & 69.31 & 57.18 \\
    \raggedright \textbf{15. Platypus2-70B} & 70.06 & 70.65 & 87.15 & 70.08 & 52.37 \\
    \bottomrule
    \end{tabular}
\end{table}

\vspace{1\baselineskip}

\begin{table}[ht]
    \centering
    \rowcolors{2}{white}{lightgray}
    \vspace{1\baselineskip}
    \caption{Top 13b open-source models according to the Hugging Face leaderboard on 8/10/23. \textbf{These rankings are for 13b parameter models only.} Our models are \nth{1}, \nth{7}, and \nth{20}. ARC-challenge is 25-shot, HellaSwag is 10-shot, MMLU is 5-shot, and TruthfulQA is 0-shot.}
    \vspace{1\baselineskip}
    \label{tab:13B_summary}
    \setlength\tabcolsep{0.5\tabcolsep}
    \begin{tabular}{@{}p{6.1cm}@{}ccccc}
    \toprule 
    \textbf{Model} & \textbf{Avg.} & \multicolumn{4}{c}{\textbf{Scores (\%)}} \\
    \cmidrule(lr){3-6}
    & & \textbf{ARC} & \textbf{HellaSwag} & \textbf{MMLU} & \textbf{TruthfulQA} \\
    \midrule
    \textbf{1. Stable-Platypus2-13B} & 63.96 & 62.71 & 82.29 & 58.3 & 52.52 \\
    \raggedright 2. Open-Orca/OpenOrcaxOpenChat-Preview2-13B & 63.83 & 62.54 & 82.96 & 58.65 & 51.17 \\
    3. psmathur/orca\_mini\_v3\_13b & 63.45 & 63.14 & 82.35 & 56.52 & 51.81 \\
    4. Gryphe/MythoMix-L2-13b & 63.11 & 61.09 & 83.86 & 55.42 & 52.08 \\
    5. stabilityai/StableBeluga-13B & 62.91 & 62.03 & 82.27 & 57.71 & 49.61 \\
    \raggedright 6. The-Face-Of-Goonery/Huginn-13b -FP16 & 62.82 & 60.58 & 82.53 & 53.71 & 54.46 \\
    \textbf{7. Camel-Platypus2-13B} & 62.62 & 60.75 & 83.61 & 56.51 & 49.6 \\
    \vdots & \vdots & \vdots & \vdots & \vdots & \vdots \\
    13. augtoma/qCammel-13B & 62.19 & 60.84 & 83.66 & 56.73 & 47.54 \\
    \vdots & \vdots & \vdots & \vdots & \vdots & \vdots \\
    \textbf{20. Platypus2-13B} & 61.35 & 61.26 & 82.56 & 56.7 & 44.86 \\
    \bottomrule
    \end{tabular}
\end{table}

The objective of our model merging strategy is to assess the synergistic effects of integrating with broad models like Instruct and Beluga, or specialized models such as Camel. An interesting observation was with the Dolphin merge, where instead of using the conventional Platypus adapters, we opted for the exported Platypus merged with the base LLaMa-2. This decision was influenced by our contamination check experiments of the Dolphin dataset. Dolphin-Platypus2-70-B is the only merge that did not do better than both the base and adapter models. Additionally, there was a smaller score discrepancy between the base Platypus and Dolphin models than the other models being discussed. This led us back to Camel, which had previously shown promising results in our initial tests using 13B. 

Post fine-tuning, both the 13B and 70B models demonstrated marked improvements over the base LLaMa-2 models, particularly in the ARC and TruthfulQA benchmarks. This prompted us to explore the potential of merging with other fine-tuned variants. While the 70B merges showed marginal variations from the baseline scores, the 13B merges, especially with Stable Beluga, displayed significant enhancements. For instance, the merge with Stable Beluga outperformed its constituent models by at least 0.5\% across most benchmarks, with a notable 2.91\% increase in TruthfulQA. Additionally, Stable-Platypus2-13B also showed an overall increase of +1.05\% jump over base model.

Given that TruthfulQA questions are primarily "knowledge" questions (as opposed to "reasoning" questions), the consistent improvement in TruthfulQA scores across merges suggests that merging models effectively broadens the knowledge base rather than enhancing reasoning capabilities. This observation aligns with the nature of TruthfulQA questions, which are primarily knowledge-based. The LLaMa-2 paper's assertion that model saturation hasn't been reached further supports the idea that merging can introduce "new" information to the model~\cite{touvron2023llama2}.

The results underscore the potential of model merging as a strategy to enhance performance. The choice of models for merging, whether broad or focused, plays a pivotal role in determining the outcome. Our experiments with Dolphin, for instance, underscore the importance of iterative testing and model selection. The consistent performance of models like Camel-Platypus2-70B across different benchmarks further emphasizes this point.

In the ARC-Challenge, Hellaswag, and TruthfulQA tests, the Camel-Platypus2-70B model exhibited the most significant positive change with a +4.12\% improvement in ARC-challenge. This suggests that the Camel-Platypus2-70B model, when merged with the Platypus adapter, is potentially the most effective combination for tasks related to the ARC-Challenge.

For the MMLU tests, the results were more varied. The Platypus2-70B-instruct model displayed a remarkable +18.18\% improvement in abstract\_algebra, while the Camel-Platypus2-13B model showed a decline of -15.62\%. This indicates that the effectiveness of the merge varies depending on the specific domain of the test. Notably, in machine\_learning, the Camel-Platypus2-70B model demonstrated a significant increase of +26.32\%, reinforcing the potential of this model in specific domains.

Drawing from the broader content of our paper, these results underscore the importance of selecting the appropriate model for merging with the Platypus adapter. The performance enhancements or declines are not uniform across all domains, emphasizing the need for domain-specific evaluations before finalizing a merge.

\subsection{Deep dive into the benchmark metric tasks}

The Appendix contains a breakdown of each MMLU task by change in percent and percent change. The rest of this discussion will be referencing percent change, but we include both for transparency.  A deeper dive into the performance metrics of the base models revealed that two models with very similar scores do not necessarily merge into a superior model.

\paragraph{ARC-Challenge, Hellaswag, TruthfulQA-MC: Table \ref{tab:p_c_3}}

\begin{itemize}
    \item Most Notable Improvement: The Camel-Platypus2-70B model in the ARC-challenge test exhibited the highest positive change with a +4.12\% improvement. This indicates that for tasks related to the ARC-Challenge, the Camel-Platypus2-70B model, when merged with the Platypus adapter, is potentially the most effective.
    \item Consistent Performer: The Stable-Platypus2-13B model showed consistent positive changes across all three tests compared to the base model, indicating its reliable performance when merged with the Platypus adapter.
    \item Variability in Results: The results for TruthfulQA were particularly varied, with the Stable-Platypus2-13B model showing a significant +5.87\% improvement, while the Dolphin-Platypus2-70B model showed a decline of -1.37\%.
\end{itemize}

\paragraph{MMLU: Table \ref{tab:p_c_mmlu})}
\begin{itemize}
    \item Standout Performance: In the machine\_learning test, the Camel-Platypus2-70B model displayed a remarkable +26.32\% improvement, indicating its potential effectiveness in machine learning domains when merged with the Platypus adapter.
    \item Diverse Results: The results for the formal\_logic test were diverse, with the Stable-Platypus2-13B model showing a significant +27.27\% improvement, while the Camel-Platypus2-13B model showed a decline of -2.13\%.
    \item Consistent Domains: In domains like marketing, the changes across all models were minimal, suggesting that the impact of merging with the Platypus adapter might be limited in certain domains.
    \item Significant Declines: The college\_physics test showed significant declines for the Platypus2-70B-instruct, Dolphin-Platypus2-70B, and Camel-Platypus2-70B models, with changes of -20.93\%, -13.16\%, and -18.42\% respectively. This indicates potential compatibility issues or inefficiencies when these models are merged with the Platypus adapter for tasks related to college physics.
\end{itemize}

The tables provide a comprehensive view of how different models perform when merged with the Platypus adapter across various domains. It's evident that the effectiveness of the merge is domain-specific, and there's no one-size-fits-all solution. Researchers and practitioners should carefully evaluate the performance enhancements or declines in their specific domain of interest before finalizing a merge.

\section{Broader Impacts \& Future Work}
Modern LLMs often require considerable computational resources, making their training and inference costs restrictive for those with limited budgets. While techniques like quantization and LoRA provide some relief, a notable observation from the Hugging Face leaderboard is the success of smaller models in specific tasks, such as role-playing and question answering. It may be strategic to harness the efficiency of these compact models and merge them with the precision of individual adapters. In that ecosystem, the similarity between inputs and training data is used as an a posteriori factor, biasing the outputs to be informed by similar data. This method essentially exploits the correlation between inputs and their similar training data to influence outputs. Mixture of Experts (MoEs) presents a promising avenue for further enhancing accuracy, given the success of domain-specific training. Future exploration could also involve integrating alpaca and orca-style datasets, as well as examining the potential of QLoRA within our pipeline.

Building on this perspective, LIMA~\cite{zhou2023lima} suggests a future characterized by an array of small, meticulously curated datasets for niche domains. The advantages of this approach are evident:
streamlined fine-tuning processes and rapid cosine similarity searches across average training inputs of adapters.

An intriguing inquiry is the applicability of the LIMA strategy within the LoRA and PEFT landscapes. This question warrants further investigation in subsequent studies. Future work might delve deeper into understanding the nuances of model merging, especially in the context of models with similar baseline scores. The potential of leveraging models like Lazarus, a successful LoRA merge of 6 models~\cite{lazarus}, could also be explored.
\section{Limitations}
Platypus, being a fine-tuned variant of LLaMa-2, inherits many of the base model's limitations while introducing some unique challenges due to its specialized training. Like LLaMa-2, Platypus does not receive continuous knowledge updates after its pretraining and fine-tuning phases. This static knowledge base can lead to outdated or incomplete information over time. Furthermore, there remains a risk of Platypus generating non-factual content or unqualified advice, especially when faced with ambiguous or misleading prompts.

While Platypus has been fine-tuned to improve its proficiency in STEM and logic, its primary focus, like LLaMa-2, has been on English-language data. Although it might exhibit some capability in other languages, this proficiency is not guaranteed and can be inconsistent due to limited non-English pretraining data. Additionally, like its predecessor, Platypus can generate potentially harmful, offensive, or biased content, especially when trained on publicly available datasets. While efforts have been made to address these issues through data cleaning, challenges persist, especially for non-English languages where comprehensive datasets might be lacking.

The capabilities of Platypus, like other AI models, can be misused for malicious purposes, such as spreading misinformation or probing sensitive topics. While our model is for non-commercial use only due to the license of the training set, we have followed Meta's Responsible Use Guide with respect to fine-tuning. We have not done any adversarial attack testing or read teaming, so before deploying any applications of Platypus, developers should perform safety testing and tuning tailored to their specific applications of the model. 

Due to its specialized training, particularly in STEM and logic questions, Platypus might exhibit limitations when faced with topics outside its primary domain of expertise. Please exercise caution—it's essential to adhere to guidelines for responsible use and consider additional fine-tuning and deployment measures to ensure optimal and safe performance.

Any users of the Platypus family should ensure that there is no contamination between the Platypus training data and any benchmark test sets not explicitly used in this paper. For example, the creators of PRM800K combined the MATH train and test sets to increase training quality. We used both the train and test sets of PRM800K during training, barring any questions that were too similar to the benchmark datasets.

All aforementioned limitations pertain to our merged model variants. Again, we deliberately chose not to merge with any models that used contaminated datasets during training. While we cannot offer absolute assurances, we proceed giving the benefit of the doubt. We'd like to stress the importance of due diligence when choosing to deploy any LLM or dataset.

Lastly, we note that keyword search and cosine similarity of sentence embeddings may not be exhaustive filtering methods. While we are confident there is no contamination in our cleaned training data, it is unlikely but not impossible that some questions slipped through the cracks. 

\section*{Acknowledgments}
A very special thank you to both Hugging Face, for creating a space where anyone can evaluate and release LLMs, and Meta AI for sharing LLaMa-2, the backbone of our fine-tuned models. We would also like to thank the creators of LoRA, without whom we could not have afforded to fine-tune a 70B variant of LLaMa-2.

\bibliographystyle{abbrv}
\bibliography{neurips}
\newpage
\section*{Appendix}
\begin{mdframed}[frametitle={Alpaca Formatting Example with Input}]
\begin{verbatim}
Below is an instruction that describes a task, paired with an input
that provides further context. Write a response that appropriately 
completes the request.

### Instruction:
{instruction}

### Input:
{input}

### Response:
\end{verbatim}
\end{mdframed}

\begin{mdframed}[frametitle={Alpaca Formatting Example without Input}]
\begin{verbatim}
Below is an instruction that describes a task. Write a response that 
appropriately completes the request.

### Instruction:
{instruction}

### Response:
\end{verbatim}
\end{mdframed}

\begin{table}[ht]
    \small
     \centering
     \caption{Percent change over "Base" Model - ARC-Challenge, Hellaswag, TruthfulQA-MC. 
     In this context, base model refers to the model on which the adapters are merged.
     \label{tab:p_c_3}}
     \rowcolors{2}{white}{lightgray}
     \begin{tabular}{llllll}
     \toprule
     \textbf{Test Name} & \textbf{Camel-P2-13B} & \textbf{Stable-P2-13B} & \textbf{P2-70B-ins} & \textbf{Dolphin-P2-70B} & \textbf{Camel-P2-70B} \\
     arc\_challenge & -0.14 & +1.10 & +1.08 & +1.10 & +4.12 \\
     hellaswag & -0.06 & +0.02 & +0.06 & -0.14 & -0.24 \\
     truthfulqa\_mc & +4.33 & +5.87 & +0.02 & -1.37 & +0.53 \\
     \bottomrule
     \end{tabular}
\end{table}
\begin{table}[ht]
    \small
    \centering
    \caption{Change in Percent over "Base" Model - ARC-Challenge, Hellaswag, TruthfulQA-MC.
    In this context, base model refers to the model on which the adapters are merged.
    \label{tab:c_in_p_3}}
    \rowcolors{2}{white}{lightgray}
    \begin{tabular}{llllll}
    \toprule
    \textbf{Test Name} & \textbf{Camel-P2-13B} & \textbf{Stable-P2-13B} & \textbf{P2-70B-ins} & \textbf{Dolphin-P2-70B} & \textbf{Camel-P2-70B} \\
    \midrule
    arc\_challenge & -0.09 & +0.68 & +0.77 & +0.77 & +2.82 \\
    hellaswag & -0.05 & +0.02 & +0.05 & -0.12 & -0.21 \\
    truthfulqa\_mc & +2.06 & +2.91 & +0.01 & -0.78 & +0.31 \\
    \bottomrule
    \end{tabular}
\end{table}

\begin{table}[ht]
    \small
    \centering
    \caption{Percent Change over "Base" Model - MMLU.
    In this context, base model refers to the model on which the adapters are merged
    \label{tab:p_c_mmlu}}
    \rowcolors{2}{white}{lightgray}
    \begin{tabular}{llllll}
    \toprule
    \textbf{Test Name} & \textbf{C-P2-13} & \textbf{S-P2-13} & \textbf{P2-70-ins} & \textbf{D-P2-70} & \textbf{C-P2-70} \\
    \midrule
    abstract\_algebra & -15.62 & -6.06 & +18.18 & -11.11 & +11.76 \\
    anatomy & -6.67 & +12.90 & -9.09 & +1.16 & 0.00 \\
    astronomy & -3.23 & +8.75 & -7.81 & -7.20 & -6.25 \\
    business\_ethics & -3.51 & +1.69 & -4.05 & +2.86 & -2.67 \\
    clinical\_knowledge & -2.52 & 0.00 & +2.06 & +0.53 & +1.05 \\
    college\_biology & +8.43 & +8.99 & +0.83 & +2.59 & -4.92 \\
    college\_chemistry & +2.56 & -2.70 & -6.12 & 0.00 & 0.00 \\
    college\_computer\_science & 0.00 & -2.17 & -3.33 & -7.02 & -10.00 \\
    college\_mathematics & +6.67 & +8.82 & +4.76 & +2.56 & +5.13 \\
    college\_medicine & -5.38 & +2.15 & +4.39 & +2.70 & +0.86 \\
    college\_physics & +3.33 & -2.94 & -20.93 & -13.16 & -18.42 \\
    computer\_security & -1.43 & -12.16 & -1.30 & -3.80 & +1.32 \\
    conceptual\_physics & +3.13 & +4.55 & -4.82 & -3.85 & 0.00 \\
    econometrics & +10.26 & +14.71 & +3.77 & +4.08 & +5.77 \\
    electrical\_engineering & -15.79 & -8.86 & -7.45 & -10.00 & -9.28 \\
    elementary\_mathematics & +6.02 & -3.10 & -3.39 & +4.22 & +0.59 \\
    formal\_logic & -2.13 & +27.27 & +13.56 & +12.07 & +22.41 \\
    global\_facts & +21.21 & +2.63 & +4.26 & -6.52 & -5.66 \\
    hs\_biology & -4.19 & -5.29 & +2.39 & +1.64 & -0.40 \\
    hs\_chemistry & -3.41 & -1.14 & -3.51 & +3.85 & +5.66 \\
    hs\_computer\_science & -8.20 & 0.00 & -1.27 & 0.00 & -3.75 \\
    hs\_european\_history & +1.80 & 0.00 & +4.32 & +2.17 & +0.72 \\
    hs\_geography & -2.70 & -0.68 & +0.58 & -5.06 & -1.74 \\
    hs\_government\_and\_politics & +8.33 & +4.40 & +1.66 & -1.67 & -1.10 \\
    hs\_macroeconomics & -4.37 & +1.34 & +1.81 & +2.61 & -1.42 \\
    hs\_mathematics & -7.69 & +15.19 & -5.81 & -10.87 & -21.51 \\
    hs\_microeconomics & -2.26 & -2.11 & +2.20 & +1.12 & +1.12 \\
    hs\_physics & -3.51 & -4.00 & +1.41 & -2.67 & -4.17 \\
    hs\_psychology & +1.42 & +4.59 & +0.41 & -0.82 & +0.61 \\
    hs\_statistics & +3.19 & +7.37 & +2.31 & +4.96 & +2.34 \\
    hs\_us\_history & +5.23 & +8.50 & -2.12 & +0.54 & -3.21 \\
    hs\_world\_history & +5.75 & +3.37 & +0.94 & +1.44 & +2.36 \\
    human\_aging & +1.40 & -4.00 & +2.26 & -1.14 & +1.15 \\
    human\_sexuality & -1.32 & -3.37 & -5.31 & -1.83 & -7.14 \\
    international\_law & +2.33 & -2.15 & +0.96 & -2.80 & +1.94 \\
    jurisprudence & -5.19 & -2.47 & +1.12 & -2.20 & 0.00 \\
    logical\_fallacies & -4.63 & -1.74 & +2.29 & 0.00 & -5.11 \\
    machine\_learning & -15.38 & -14.00 & +22.81 & +16.07 & +26.32 \\
    management & -2.63 & -1.27 & +2.35 & 0.00 & +3.53 \\
    marketing & +1.08 & -2.58 & +0.95 & +0.94 & +0.94 \\
    medical\_genetics & +13.21 & -5.97 & 0.00 & -1.39 & -1.45 \\
    miscellaneous & +1.86 & +0.66 & +0.15 & -0.29 & -0.59 \\
    moral\_disputes & +1.81 & -0.45 & -2.96 & -1.15 & -5.04 \\
    moral\_scenarios & +3.54 & +19.74 & +7.95 & +17.71 & +6.37 \\
    nutrition & -5.43 & 0.00 & -2.98 & +2.23 & -2.54 \\
    philosophy & +1.00 & +2.45 & 0.00 & +1.25 & +1.25 \\
    prehistory & +1.46 & +6.83 & 0.00 & +3.01 & -1.47 \\
    professional\_accounting & +10.00 & +4.10 & -1.23 & +3.29 & -1.90 \\
    professional\_law & +8.01 & +10.05 & +6.61 & +5.31 & +5.13 \\
    professional\_medicine & +4.29 & +9.59 & -1.49 & -2.50 & -3.40 \\
    professional\_psychology & +4.69 & +3.64 & -1.07 & +0.22 & +0.22 \\
    public\_relations & -5.33 & +5.71 & -4.88 & -1.25 & 0.00 \\
    security\_studies & -2.03 & -3.16 & -5.47 & -3.08 & -0.52 \\
    sociology & -5.92 & -6.16 & +1.14 & +1.14 & +0.58 \\
    us\_foreign\_policy & -8.54 & -4.82 & -4.44 & -4.40 & -3.33 \\
    virology & -5.41 & -1.28 & +1.14 & -2.20 & +4.60 \\
    world\_religions & +0.75 & +0.75 & -2.00 & -2.03 & -3.29 \\
    \bottomrule
    \end{tabular}
    \vspace{1em} 
    
    \begin{minipage}{\textwidth}
        \textit{Note:} C-P2-13 refers to Camel-Platypus2-13B, S-P2-13 refers to Stable-Platypus2-13B, D-P2-70 refers to Dolphin-Platypus2-70B, and C-P2-70 refers to Camel-Platypus2-70B.
    \end{minipage}
\end{table}

\begin{table}[ht]
    \small
    \centering
    \caption{Change in Percent over "Base" Model - MMLU.
    In this context, base model refers to the model on which the adapters are merged.
    \label{tab:c_in_p_mmlu}}
    \rowcolors{2}{white}{lightgray}
    \begin{tabular}{llllll}
    \toprule
    \textbf{Test Name} & \textbf{C-P2-13} & \textbf{S-P2-13} & \textbf{P2-70-ins} & \textbf{D-P2-70} & \textbf{C-P2-70} \\
    \midrule
    abstract\_algebra & -5.00 & -2.00 & +6.00 & -4.00 & +4.00 \\
    anatomy & -3.70 & +5.93 & -5.93 & +0.74 & 0.00 \\
    astronomy & -1.97 & +4.61 & -6.58 & -5.92 & -5.26 \\
    business\_ethics & -2.00 & +1.00 & -3.00 & +2.00 & -2.00 \\
    clinical\_knowledge & -1.51 & 0.00 & +1.51 & +0.38 & +0.75 \\
    college\_biology & +4.86 & +5.56 & +0.69 & +2.08 & -4.17 \\
    college\_chemistry & +1.00 & -1.00 & -3.00 & 0.00 & 0.00 \\
    college\_computer\_science & 0.00 & -1.00 & -2.00 & -4.00 & -6.00 \\
    college\_mathematics & +2.00 & +3.00 & +2.00 & +1.00 & +2.00 \\
    college\_medicine & -2.89 & +1.16 & +2.89 & +1.73 & +0.58 \\
    college\_physics & +0.98 & -0.98 & -8.82 & -4.90 & -6.86 \\
    computer\_security & -1.00 & -9.00 & -1.00 & -3.00 & +1.00 \\
    conceptual\_physics & +1.28 & +2.13 & -3.40 & -2.55 & 0.00 \\
    econometrics & +3.51 & +4.39 & +1.75 & +1.75 & +2.63 \\
    electrical\_engineering & -8.28 & -4.83 & -4.83 & -6.21 & -6.21 \\
    elementary\_mathematics & +2.12 & -1.06 & -1.59 & +1.85 & +0.26 \\
    formal\_logic & -0.79 & +9.52 & +6.35 & +5.56 & +10.32 \\
    global\_facts & +7.00 & +1.00 & +2.00 & -3.00 & -3.00 \\
    hs\_biology & -2.90 & -3.55 & +1.94 & +1.29 & -0.32 \\
    hs\_chemistry & -1.48 & -0.49 & -1.97 & +1.97 & +2.96 \\
    hs\_computer\_science & -5.00 & 0.00 & -1.00 & 0.00 & -3.00 \\
    hs\_european\_history & +1.21 & 0.00 & +3.64 & +1.82 & +0.61 \\
    hs\_geography & -2.02 & -0.51 & +0.51 & -4.55 & -1.52 \\
    hs\_government\_and\_politics & +6.74 & +3.63 & +1.55 & -1.55 & -1.04 \\
    hs\_macroeconomics & -2.56 & +0.77 & +1.28 & +1.79 & -1.03 \\
    hs\_mathematics & -2.59 & +4.44 & -1.85 & -3.70 & -7.41 \\
    hs\_microeconomics & -1.26 & -1.26 & +1.68 & +0.84 & +0.84 \\
    hs\_physics & -1.32 & -1.32 & +0.66 & -1.32 & -1.99 \\
    hs\_psychology & +1.10 & +3.49 & +0.37 & -0.73 & +0.55 \\
    hs\_statistics & +1.39 & +3.24 & +1.39 & +2.78 & +1.39 \\
    hs\_us\_history & +3.92 & +6.37 & -1.96 & +0.49 & -2.94 \\
    hs\_world\_history & +4.22 & +2.53 & +0.84 & +1.27 & +2.11 \\
    human\_aging & +0.90 & -2.69 & +1.79 & -0.90 & +0.90 \\
    human\_sexuality & -0.76 & -2.29 & -4.58 & -1.53 & -6.11 \\
    international\_law & +1.65 & -1.65 & +0.83 & -2.48 & +1.65 \\
    jurisprudence & -3.70 & -1.85 & +0.93 & -1.85 & 0.00 \\
    logical\_fallacies & -3.07 & -1.23 & +1.84 & 0.00 & -4.29 \\
    machine\_learning & -5.36 & -6.25 & +11.61 & +8.04 & +13.39 \\
    management & -1.94 & -0.97 & +1.94 & 0.00 & +2.91 \\
    marketing & +0.85 & -2.14 & +0.85 & +0.85 & +0.85 \\
    medical\_genetics & +7.00 & -4.00 & 0.00 & -1.00 & -1.00 \\
    miscellaneous & +1.40 & +0.51 & +0.13 & -0.26 & -0.51 \\
    moral\_disputes & +1.16 & -0.29 & -2.31 & -0.87 & -4.05 \\
    moral\_scenarios & +1.56 & +8.60 & +4.80 & +9.50 & +3.58 \\
    nutrition & -3.27 & 0.00 & -2.29 & +1.63 & -1.96 \\
    philosophy & +0.64 & +1.61 & 0.00 & +0.96 & +0.96 \\
    prehistory & +0.93 & +4.32 & 0.00 & +2.47 & -1.23 \\
    professional\_accounting & +4.26 & +1.77 & -0.71 & +1.77 & -1.06 \\
    professional\_law & +3.46 & +4.17 & +3.65 & +2.87 & +2.87 \\
    professional\_medicine & +2.57 & +5.15 & -1.10 & -1.84 & -2.57 \\
    professional\_psychology & +2.61 & +2.12 & -0.82 & +0.16 & +0.16 \\
    public\_relations & -3.64 & +3.64 & -3.64 & -0.91 & 0.00 \\
    security\_studies & -1.22 & -2.04 & -4.49 & -2.45 & -0.41 \\
    sociology & -4.48 & -4.48 & +1.00 & +1.00 & +0.50 \\
    us\_foreign\_policy & -7.00 & -4.00 & -4.00 & -4.00 & -3.00 \\
    virology & -2.41 & -0.60 & +0.60 & -1.20 & +2.41 \\
    world\_religions & +0.58 & +0.58 & -1.75 & -1.75 & -2.92 \\
    \bottomrule
    \end{tabular}
    \vspace{1em} 
    
    \begin{minipage}{\textwidth}
        \textit{Note:} C-P2-13 refers to Camel-Platypus2-13B, S-P2-13 refers to Stable-Platypus2-13B, D-P2-70 refers to Dolphin-Platypus2-70B, and C-P2-70 refers to Camel-Platypus2-70B.
    \end{minipage}
\end{table}

\begin{table}
    \centering
    \caption{Hyperparameters for 13B and 70B Models}
    \vspace{1\baselineskip}
    \label{tab:hyperparams}
    \begin{tabular}{ll}
    \toprule
    \textbf{Hyperparameter} & \textbf{Platypus2-13B / 70B} \\
    \midrule
    batch size & 16 \\
    micro batch size & 1 \\
    num epochs & 1 \\
    learning rate & 4e-4 / 3e-4 \\
    cutoff len & 4096 \\
    lora rank & 16 \\
    lora alpha & 16 \\
    lora dropout & 0.05 \\
    lora target modules & gate\_proj, down\_proj, up\_proj \\
    train on inputs & False \\
    add eos token & False \\
    group by length & False \\
    prompt template & alpaca \\
    lr scheduler & cosine \\
    warmup steps & 100 \\
    \bottomrule
    \end{tabular}
\end{table}

\end{document}